%% file: acl2023.tex
\pdfoutput=1

\documentclass[11pt]{article}

\usepackage{ACL2023}

\usepackage{times}
\usepackage{latexsym}

\usepackage[T1]{fontenc}

\usepackage[utf8]{inputenc}

\usepackage{microtype}

\usepackage{inconsolata}

\usepackage{graphicx}
\usepackage{tabularx}
\usepackage{booktabs}
\usepackage{multirow}
\usepackage{tikz}
\PassOptionsToPackage{hyphens}{url}\usepackage{hyperref}
\usepackage{xurl}
\title{A Weakly Supervised Classifier and Dataset of \\ White Supremacist Language}

\author{Michael Miller Yoder\textmd{\textsuperscript{1}} \hspace*{1ex} Ahmad Diab\textmd{\textsuperscript{2}} \hspace*{1ex} David West Brown\textmd{\textsuperscript{3}} \hspace*{1ex} Kathleen M. Carley\textmd{\textsuperscript{1}} \\
    \textsuperscript{1}Software and Societal Systems Department, Carnegie Mellon University \\
    \textsuperscript{2}Department of Computer Science, University of Pittsburgh \\
    \textsuperscript{3}Department of English, Carnegie Mellon University \\
    Pittsburgh, Pennsylvania, USA \\
    \texttt{yoder@cs.cmu.edu} \hspace*{1ex} \texttt{ahd23@pitt.edu} \hspace*{1ex} \texttt{dwb2@andrew.cmu.edu} \hspace*{1ex} \texttt{carley@cs.cmu.edu} \\
}
\begin{document}
\maketitle
\begin{abstract}
We present a dataset and classifier for detecting the language of white supremacist extremism, a growing issue in online hate speech.
Our weakly supervised classifier is trained on large datasets of text from explicitly white supremacist domains paired with neutral and anti-racist data from similar domains.
We demonstrate that this approach improves generalization performance to new domains. 
Incorporating anti-racist texts as counterexamples to white supremacist language mitigates bias.
\end{abstract}

\section{Introduction}
The spread of white supremacist extremism online has motivated offline violence, including recent mass shootings in Christchurch, El Paso, Pittsburgh, and Buffalo. 
Though some research in natural language processing has focused on types of hate speech, such as anti-Black racism~\citep{Kwok2013} and misogyny~\citep{fersini_overview_2018}, little work has focused on detecting specific hateful ideologies.
Practitioners have called for such systems, particularly for white supremacism~\citep{anti-defamation_league_deplatform_2022, yoder_research_2022}.

To detect white supremacist language, we build text classifiers trained on data from a large, diverse set of explicitly white supremacist online spaces, filtered to ideological topics.\footnote{See \url{https://osf.io/274z3/} to access public parts of this dataset and others used in this paper.}
In a weakly supervised set-up, we train discriminative classifiers to distinguish texts in white supremacist domains from texts in similar online spaces that are not known for white supremacism.
These classifiers outperform prior work in white supremacist classification on three annotated datasets, and we find that the best-performing models use a combination of weakly and manually annotated data.

Hate speech classifiers often have difficulty generalizing beyond data they were trained on~\citep{swamy_studying_2019,yoder_how_2022}.
We evaluate our classifiers on unseen datasets annotated for white supremacism from a variety of domains and find strong generalization performance for models that incorporate weakly annotated data.

Hate speech classifiers often learn to associate any mention of marginalized identities with hate, regardless of context~\citep{Dixon2017}.
To address this potential issue with white supremacist classification, we incorporate anti-racist texts, which often mention marginalized identities in positive contexts, as counter-examples to white supremacist texts.
Evaluating on a synthetic test set with mentions of marginalized identities in a variety of contexts~\citep{rottger_hatecheck_2021}, we find that including anti-racist texts helps mitigate this bias.

\section{The Language of White Supremacist Extremism}
\label{sec:background}
This work focuses on white supremacist extremism, social movements advocating for the superiority of white people and domination or separation from other races \citep{daniels2009cyber}.
This fringe movement both exploits the bigotries widely held in societies with structural white supremacism and makes them explicit~\citep{ferber2004home,berlet_overview_2006,pruden_birds_2022}.
Key beliefs of white supremacist extremism are that race and gender hierarchies are fixed, that white people's ``natural'' power is threatened, and that action is needed to protect the white race~\citep{ferber_reading_2000,brown_wwwhatecom_2009,perry_white_2016,ansah_violent_2021}.

Many qualitative studies have examined the language of white supremacism \citep{thompson_watching_2001,duffy_web_2003,perry_white_2016,bouvier_covert_2020}.
Computational models have been developed to identify affect~\citep{figea_measuring_2016}, hate speech~\citep{DeGibert2019}, and violent intent~\citep{simons_bootstrapped_2020} within white supremacist forums.

Two other studies have built models to detect white supremacist ideology in text.
\citet{alatawi_detecting_2021} test Word2vec/BiLSTM models, pre-trained on a corpus of unlabeled white supremacist forum data, as well as BERT models.
To estimate the prevalence of white supremacism on Twitter after the 2016 US election, \citet{siegel_trumping_2021} build a dictionary-based classifier and validate their findings with unlabeled alt-right Reddit data.
In contrast, we use a large, domain-general white supremacist corpus with carefully selected negative training examples to build a weakly supervised discriminative classifier for white supremacism.

\subsection{Hate speech and white supremacism}
The relationship between hate speech and white supremacism has been theorized and annotated in different ways.
Some have annotated the glorification of ideologies and groups such as Nazism and the Ku Klux Klan separately from hate speech~\citep{siegel_trumping_2021,rieger_assessing_2021}, which is often defined as verbal attacks on groups based on their identity~\citep{sanguinetti_italian_2018,poletto_resources_2021,DeGibert2019}.
A user of Stormfront, a white supremacist forum, notes this distinction to evade moderation on other platforms: ``Nationalist means defending the white race; racist means degrading non-white races. You should be fine posting about preserving the white race as long as you don't degrade other races.''\footnote{Quotes in this paper are paraphrased for privacy~\citep{williams_towards_2017}}

We aim to capture the expression of white supremacist ideology beyond just hate speech against marginalized identities  (see Figure \ref{fig:venn_diagram}).
In contrast, \citet{DeGibert2019} ask annotators to identify hate speech within a white supremacist forum.
They note that some content that did not fit strict definitions of hate speech still exhibited white supremacist ideology. 
Examples of this from data used in the current paper include ``diversity means chasing down whites'' (white people being threatened) and ``god will punish as he did w/ hitler'' (action needed to protect white people).

\input{tikz/venn_diagram}

\begin{table*}[tb]
\begin{tabularx}{\textwidth}{| p{98pt} | p{88pt} | r | X |}
\hline    
Data source & Platform & \# Posts & Excerpt from example post \\
\hline    
\citet{papasavva_raiders_2020} & 4chan & 2,686,267 & africans are inferior animals \\
Stormfront archive & Stormfront & 751,980 & help the white race \\
\citet{jokubauskaite_generally_2020} & 4chan & 578,650 & we need to drop the nazism no , we need to do the opposite \\
Iron March archive & Iron March & 179,468 & disgusting looking fat ch*nk cuckold \\
\citet{qian_hierarchical_2018} & Twitter & 84,695 & keep illegal immigrants out \\
Patriot Front archive & Discord & 39,577 & interracial dating i find that appalling \\
\citet{calderon_linguistic_2021} & \begin{tabular}[t] {@{}l@{}} Daily Stormer, \\ Amer. Renaissance \end{tabular} & 26,099 &  black - on - white murders  it never ends \\
\citet{pruden_birds_2022} & books, manifestos & 17,007 &  preventing the ongoing islamisation \\
\citet{elsherief-etal-2021-latent} & Twitter & 3,480 & desert barbarians will destroy the west \\
\hline
\end{tabularx}
\caption{Information on white supremacist corpus before filtering and sampling. \textit{\textbf{Warning: offensive examples.}}}
\label{tab:ws_data_full}
\end{table*}

\section{Weakly Annotated Data}
It is difficult for annotators to determine whether the short texts commonly used in NLP and computational social science, such as tweets, express white supremacism or other far-right ideologies.
\citet{alatawi_detecting_2021} struggle to reach adequate inter-annotator agreement on white supremacism in tweets.
\citet{hartung_identifying_2017} note that individual tweets are difficult to link to extreme right-wing ideologies and instead choose to annotate user tweet histories.

Instead of focusing on individual posts, we turn to \textit{weak supervision}, approaches to quickly and cheaply label large amounts of training data based on rules, knowledge bases or other domain knowledge~\citep{ratner_snorkel_2017}.
Weakly supervised learning has been used in NLP for tasks such as cyberbullying detection~\citep{raisi_cyberbullying_2017}, sentiment analysis~\citep{kamila_ax-mabsa_2022}, dialogue systems~\citep{hudecek_discovering_2021} and others~\citep{karamanolakis-etal-2021-self}.
For training the discriminative white supremacist classifier, we draw on three sources of text data with ``natural'' (weak) labels: white supremacist domains and organizations, neutral data with similar topics, and anti-racist blogs and organizations.

\subsection{White supremacist data}
\label{sec:ws_data}
We sample existing text datasets and data archives from white supremacist domains and organizations to build a dataset of texts that likely express white supremacist extremism.
\autoref{tab:ws_data_full} details information on source datasets.

Sources include sites dedicated to white supremacism, such as Stormfront, Iron March, and the Daily Stormer.
When possible, we filter out non-ideological content on these forums using existing topic structures, for example, excluding ``Computer Talk'' and ``Opposing Views'' forums on Stormfront.
We also include tweets from organizations that the Southern Poverty Law Center labels as white supremacist hate groups~\citep{qian_hierarchical_2018,elsherief-etal-2021-latent}.
In \citeposs{papasavva_raiders_2020} dataset from the 4chan /pol/ ``politically incorrect'' imageboard, we select posts from users choosing Nazi, Confederate, fascist, and white supremacist flags.
We also include 4chan /pol/ posts in ``general'' threads with fascist and white supremacist topics \citep{jokubauskaite_generally_2020}.
From \citet{pruden_birds_2022}, we include white supremacist books and manifestos.
We also include leaked chats from Patriot Front, a white supremacist group.
Details on these datasets can be found in Appendix \ref{app:ws_corpus}. 

With over 230 million words in 4.3 million posts across many domains, this is the largest collection of white supremacist text we are aware of.
Contents are from 1968 through 2019, though 76\% of posts are from 2017-2019 (see distributions of posts over time in Appendix \ref{app:ws_corpus}).

\paragraph{Outlier filtering and sampling}
This large dataset from white supremacist domains inevitably contains many posts that are off-topic and non-ideological.
To build a weakly supervised classifier, we wish to further filter to highly ideological posts from a variety of domains.

We first remove posts with 10 or fewer words, as these are often non-ideological or require context to be understood (such as ``reddit and twitter are cracking down today'' or ``poor alex, i feel bad'').

We then select posts whose highest probability topic from an LDA model~\citep{Blei2003} are ones that are more likely to express white supremacist ideology. 
LDA with 30 topics separated themes well based on manual inspection. 
One of the authors annotated 20 posts from each topic for expressing a tenet of white supremacism, described in Section \ref{sec:background}.
We selected 6 topics with the highest annotation score for white supremacy, as this gave the best performance on evaluation datasets.
These topics related to antisemitism, anti-Black racism, and discussions of European politics and Nazism (details in Appendix \ref{app:lda}).
To balance forum posts with other domains and approximate domain distributions in neutral and anti-racist datasets, we randomly sample 100,000 forum posts.
This white supremacist corpus used in experiments contains 118,842 posts and 10.7 million words.

\subsection{Neutral data}
We also construct a corpus of ``neutral'' (not white supremacist) data that matches the topics and domains of the white supremacist corpus.
To match forum posts, we sample r/politics and r/Europe subreddits.
To match tweets, we query the Twitter API by sampling the word distribution in white supremacist tweets after removing derogatory language.
For articles, we sample random US news from the News on the Web (NOW) Corpus\footnote{\url{https://www.corpusdata.org/now_corpus.asp}}, and use a random Discord dataset to match chat~\citep{discord-data}.
For each of these domains, we sample the same number of posts per year as is present in the white supremacist corpus.
If there is not significant time overlap, we sample enough posts to reach a similar word count.
This corpus contains 159,019 posts and 8.6 million words.

\subsection{Anti-racist data}
Hate speech classifiers often overpredict mentions of marginalized identities as hate~\citep{Dixon2017}.
Assuming our data is biased until proven innocent~\citep{hutchinson_towards_2021}, we design for this issue.
We hypothesize that texts from anti-racist perspectives may help.
Oxford Languages defines anti-racism as movements ``opposing racism and promoting racial equality''.
Anti-racist communications often mention marginalized identities (as do white supremacist texts), but cast them in positive contexts, such as a tweet in our anti-racist dataset that reads, ``stand up for \#immigrants''.

We construct a corpus of anti-racist texts to match the domain and year distribution of the white supremacist corpus.
For forum data, we sample comments in subreddits known for anti-racism: r/racism, r/BlackLivesMatter, and r/StopAntiAsianRacism.
We include tweets from anti-racist organizations listed by the University of North Carolina Diversity and Inclusion office\footnote{\url{https://diversity.unc.edu/anti-racism-resources/}}.
To match articles, we scrape Medium blog posts tagged with ``anti-racism'', ``white supremacy'', ``racism'', and ``BlackLivesMatter''.
As with other corpora, data from each of these sources was inspected for its perspective.
This anti-racist corpus contains 87,807 posts and 5.6 million words.

\section{Classification}
Due to the success of BERT-based hate speech models~\citep{mozafari_bert-based_2019,samghabadi_aggression_2020}, we select the parameter-efficient DistilBERT model~\citep{sanh_distilbert_2019} to compare data configurations\footnote{Code for experiments and dataset processing is available at \url{https://github.com/michaelmilleryoder/white_supremacist_lang}.}.
We use a learning rate of $2 \times 10^{-5}$, batch size of 16, and select the epoch with the highest ROC AUC on a 10\% development set, up to 5 epochs.
Training each model took approximately 8 hours on an NVIDIA RTX A6000 GPU.

We train models on binary white supremacist classification.
All posts in the white supremacist corpus, after sampling and filtering, are labeled `white supremacist'. 
Posts in neutral and anti-racist corpora are labeled `not white supremacist'.
We also test combining weakly labeled data with manually annotated data from existing datasets (see below) and our own annotation of white supremacist posts in LDA topics.
Since there is relatively little manually annotated data, we duplicate it 5 times in these cases, to a size of 57,645 posts. 

\subsection{Evaluation}
Evaluating weakly supervised classifiers on a held-out weakly supervised set may overestimate performance.
Classifiers may learn the idiosyncrasies of domains known for white supremacy in contrast to neutral domains (4chan vs. Reddit, e.g.) instead of learning distinctive features of white supremacy.
We thus evaluate classifiers on their ability to distinguish posts manually annotated for white supremacy within the same domains, in the following 3 datasets:

\textbf{\citet{alatawi_detecting_2021}}: 1100 out of 1999 tweets (55.0\%) annotated as white supremacist.
Like our work, they conceptualize white supremacy as including hate speech against marginalized groups.

\textbf{\citet{rieger_assessing_2021}}: 366 out of 5141 posts (7.1\%) from 4chan, 8chan, and r/the\_Donald annotated as white supremacist.
This work uses a more restricted definition of white supremacy largely distinct from hate speech.
We sample examples labeled as white supremacist or neither white supremacist nor hate speech.
Examples only annotated as hate speech are excluded since they may or may not fit our broader conception of white supremacism.

\textbf{\citet{siegel_trumping_2021}}: 171 out of 9743 tweets (1.8\%) annotated as white supremacist.
Since they use a more restrictive definition of white supremacy, we sample posts annotated as white supremacist or neither white supremacist nor hate speech.

The proportions of white supremacist posts in these annotated evaluation datasets vary widely, so we report ROC AUC instead of precision, recall, or F1-score, which assume similar class proportions between training and test data~\citep{ma2013imbalanced}.
Precision and recall curves are also available in Figure \ref{fig:pr_curves} in Appendix \ref{app:eval_datasets}.

\paragraph{Generalization evaluation}
To test the ability of classifiers to generalize, we perform a leave-one-out test among annotated datasets.
During three runs for each model that uses manually annotated data, we train on two of the annotated datasets and test performance on the third.
To test generalization to a completely unseen domain, we use a dataset of quotes from offline white supremacist propaganda, extracted from data collected by the Anti-Defamation League (ADL)\footnote{\url{https://www.adl.org/resources/tools-to-track-hate/heat-map}}.
1655 out of 1798 quotes (92.0\%) were annotated by two of the authors as exhibiting white supremacist ideology.

\paragraph{Baselines}
We evaluate our approaches against the best-performing model from \citet{alatawi_detecting_2021}, BERT trained on their annotated Twitter dataset for 3 epochs with a learning rate of $2 \times 10^{-5}$ and batch size of 16.
We also compare against \citet{siegel_trumping_2021}, who first match posts with a dictionary and then filter out false positives with a Naive Bayes classifier.
Though \citet{rieger_assessing_2021} also present data annotated for white supremacy, they focus on analysis and do not propose a classifier.

\paragraph{HateCheck evaluation for lexical bias}
To evaluate bias against mentions of marginalized identities, we use the synthetic HateCheck dataset~\citep{rottger_hatecheck_2021}.
We filter to marginalized racial, ethnic, gender and sexual identities, since white supremacy is a white male perspective interlinked with misogyny and homophobia~\citep{ferber2004home,brindle2016}.
We select sentences that include these identity terms in non-hateful contexts: neutral and positive uses; homonyms and reclaimed slurs; and counterspeech of quoted, referenced, and negated hate speech.
This sample totals 762 sentences.

\section{Results}
\autoref{tab:train_test_results} presents performance of single runs on randomly sampled 30\% test sets from \citet{alatawi_detecting_2021}, \citet{rieger_assessing_2021}, and \citet{siegel_trumping_2021}.
Classifiers trained with both weakly annotated data and a combination of all manually annotated data average the best performance across evaluation datasets.
On the \citet{alatawi_detecting_2021} dataset, their own classifier performs the best.
All models have lower scores on this challenging dataset, which human annotators also struggled to agree on (0.11 Cohen's $\kappa$).
In generalization performance (\autoref{tab:generalization}), we find that using weakly annotated data outperforms using only manually annotated data in almost all cases, and that combining weakly and manually annotated data enables classifiers to generalize most effectively.

\begin{table}[tb]
\centering
\begin{tabular}{lrrrr}
\toprule
Model & \textsf{A} & \textsf{R} & \textsf{S} & Mean \\
\midrule
\textsf{S} & 60.3 & 61.8 & 61.3 & 61.2 \\
\textsf{A} & \textbf{74.0} & 81.2 & 89.7 & 81.6 \\
Annotated & 65.3 & 86.1 & 92.9 & 81.4 \\
Weak & 71.6 & 87.8 & 90.3 & 83.2 \\
Weak + Ann & 70.9 & \textbf{90.3} & \textbf{96.8} & \textbf{86.0} \\
\bottomrule
\end{tabular}
\caption{ROC AUC scores of models (rows) on test splits of evaluation datasets (columns).
\textsf{A} = \citet{alatawi_detecting_2021}, \textsf{R} = \citet{rieger_assessing_2021}, \textsf{S} = \citet{siegel_trumping_2021}.
}
\label{tab:train_test_results}
\end{table}

\begin{table}[tb]
\centering
\begin{tabular}{lrrr|r}
\toprule
Model & \textsf{A} & \textsf{R} & \textsf{S} & ADL \\
\midrule
\textsf{S} & 56.3 & 61.9 & - & 57.2 \\
\textsf{A} & - & 81.9 & 83.9 & 89.1 \\
Annotated & 55.2 & 82.0 & 84.7 & 68.5 \\
Weak & \textbf{71.0} & 87.8 & 87.3 & 85.1 \\
Weak + Ann & 70.0 & \textbf{89.8} & \textbf{88.9} & \textbf{89.2} \\
\bottomrule
\end{tabular}
\caption{Generalization performance (ROC AUC). 
For Annotated and Weak + Annotated models, the first 3 columns report scores on evaluation datasets when trained on data from the other two datasets.
The final column reports on the unseen ADL dataset.
Scores on datasets used by baseline models for training are not reported since this table focuses on generalization.
}
\label{tab:generalization}
\end{table}

\subsection{Anti-racist corpus}
Training with both neutral and anti-racist negative examples improves accuracy on the HateCheck dataset to 69.2 from 60.5 when using a similar number of only neutral negative examples.
This supports our hypothesis that incorporating anti-racist texts can mitigate bias against marginalized identity mentions.
Adding anti-racist texts slightly decreases performance on the other 4 evaluation datasets, to 82.8 from 84.3 mean ROC AUC.

\section{Conclusion}
Ideologies such as white supremacy are difficult to annotate and detect from short texts.
We use weakly supervised data from domains known for white supremacist ideology to develop classifiers that outperform and generalize more effectively than prior work.
Incorporating texts from an anti-racist perspective mitigates lexical bias.

To apply a white supremacist language classifier to varied domains, our results show the benefit of using such weakly supervised data, especially in combination with a small amount of annotated data.
Other methods for combining these data could be explored in future work, such as approaches that use reinforcement learning to select unlabeled data for training~\citep{ye_zero-shot_2020,pujari-etal-2022-reinforcement}.
Incorporating social science insights and looking for specific tenets of white supremacist extremism could also lead to better classification.
This classifier could be applied to measure the prevalence or spread of white supremacist ideology through online social networks.



\section*{Limitations}
The presented classifier and dataset are only from English-speaking sources, a major disadvantage in detecting white supremacist content globally.
The dataset also is predominantly sourced from data between 2015-2019 and reflects white supremacist extremist responses to current events from that period, including the Black Lives Matter movement.
This limits its effectiveness in detecting white supremacist content from other time periods.

Though including anti-racist data helps mitigate bias tested by our sample of the HateCheck dataset, an accuracy of 69.2\% shows room for improvement.
There is still a risk of overclassifying posts with marginalized identity mentions as white supremacist.

\section*{Ethics Statement}
There are significant ethical issues to consider in developing text classifiers for ideologies.
Since this research has clear social implications, we wish to be explicit about values and author positionality beyond a sense of ``objectivity'' in selecting research questions~\cite{schlesinger_intersectional_2017,dignazio_data_2020,waseem_disembodied_2021}.
The authors come from European- and American-dominated university contexts and consider working against racism and white supremacy a priority.
Most identify as white and some identify as people of color.
This research proceeded with values of racial justice and places those values at the center of assessing knowledge claims~\citep{collins1990black,daniels2009cyber}.
Our choice of focusing on white supremacy among other ideologies stems from those values.
White supremacist extremism, as well as structural white supremacism, is responsible for substantial harms against those with marginalized identities.
This research responds to a need from practitioners for more nuanced classifiers than for broad categories of hate speech or abusive language.
We thus choose to pursue this research, though caution that developing classifiers for other ideologies should be done with careful consideration and a clear statement of motivating values.

There are significant risks which we consider, and attempt to mitigate, in such a dataset and classifier.
First, there is the risk of misuse of a large corpus of white supremacist data, as has been seen in building and releasing a hate speech ``troll bot'' from 4chan data\footnote{\url{https://www.vice.com/en/article/7k8zwx/ai-trained-on-4chan-becomes-hate-speech-machine}}.
For this reason we build a discriminative, not generative, classifier, and only plan on releasing our dataset through a vetting process instead of publicly.

There are also privacy risks in how such a classifier could be used.
Our classifier only identifies language that is likely similar to white supremacist content.
The intended use of this classifier is to measure the prevalence of such an ideology on particular platforms or within networks for research purposes, not to label individuals as holding or not holding white supremacist ideologies.
Using the classifier for this purpose poses significant risks of misclassification and could increase harmful surveillance tactics.
We strongly discourage such a use.
Our hope is that our proposed classifier and dataset can increase knowledge about the nature and extent of white supremacist extremist movement online and can inform structural interventions, such as platform policies, not interventions against individuals.

Hate speech classifiers, developed by researchers with similar equity-based values, have been found to contain biases against marginalized groups \citep{sap_risk_2019, davidson_racial_2019}.
We measure and mitigate this bias from the start by incorporating anti-racist data, though caution that this risk still exists.

\section*{Acknowledgements}
This work was supported in part by the Collaboratory Against Hate: Research and Action Center at Carnegie Mellon University and the University of Pittsburgh. 
The Center for Informed Democracy and Social Cybersecurity at Carnegie Mellon University also provided support.
We thank the researchers who provided source datasets, including Diana Rieger, Alexandra Siegel and others at the Center for Social Media and Politics at New York University, Jherez Taylor, Jing Qian, and Meredith Pruden.
We also thank the Internet Archive and investigations teams at Bellingcat and Unicorn Riot for archiving source datasets online, and Maarten Sap for feedback.

\bibliography{anthology,custom,zotero_references}
\bibliographystyle{acl_natbib}

\appendix
\section{White supremacist corpus details}
\label{app:ws_corpus}
We sample 9 datasets and data dumps to construct our white supremacist corpus (see Section \ref{sec:ws_data}).
Here we provide details on how each of these data sources was processed and sampled, as well as other details of the corpus.

\paragraph{\citet{papasavva_raiders_2020}:}
    4chan /pol/ allows users to select ``troll'' flags to use instead of the default country flag detected from their IP address.
    We filter this dataset\footnote{Available at \url{https://zenodo.org/record/3606810\#.Y8lkkBXMKF6}, accessed 19 January 2023. This dataset is under a Creative Commons Attribution 4.0 International license.} 
    to posts from users that chose to post with Nazi, White Supremacist, Confederate, or Fascist troll flags.
    From a qualitative check, samples of posts from users with these flags often expressed white supremacist ideology.
    We remove posts with duplicate texts, as well as posts that are also found in the 4chan /pol/ dump from \citet{jokubauskaite_generally_2020}.
    Our sample of this dataset contains posts from 2017 through 2019.

    \paragraph{Stormfront data archive:} Stormfront, a popular white supremacist forum, is no longer active.
    We sample from an Internet Archive dump of its content taken in 2017\footnote{Available at \url{https://archive.org/details/stormfront.org_201708}, accessed 11 January 2023}.
    We extract forum text from the HTML files and exclude threads that are not in English and are non-ideological.
    Specifically, we exclude the following threads: Nederland \& Vlaanderen, Srbija, Español y Portugués,  Italia, Croatia, South Africa, en Français, Russia, Baltic / Scandinavia, Hungary, Opposing Views Forum, Computer Talk.
    Our sample of this dataset contains posts from 2001 through 2017.
    
\paragraph{\citet{jokubauskaite_generally_2020}:}
    We select posts in this dataset of ``general'' 4chan /pol/ threads\footnote{Available at \url{https://zenodo.org/record/3603292\#.Y8lmTxXMKF5}, accessed 19 January 2023. This dataset is under a Creative Commons Attribution 4.0 International license.} that we find to be related to white supremacy and fascism: kraut/pol/, afd, national socialism, fascism, dixie, kraut/pol/, ethnostate, white, chimpout, feminist apocalypse, (((krautgate))).
    This dataset contains posts from 2001 through 2017.

\paragraph{Iron March data archive:}
    Data from Iron March, a now defunct neo-Nazi and white supremacist message board, was obtained through an Internet Archive data dump\footnote{Available through links at \url{https://www.bellingcat.com/resources/how-tos/2019/11/06/massive-white-supremacist-message-board-leak-how-to-access-and-interpret-the-data/}, accessed 11 January 2023} referenced in \citet{simons_bootstrapped_2020}.
    This dataset contains posts from 2011 through 2017.

\paragraph{\citet{qian_hierarchical_2018}:} We rehydrate tweet IDs from this dataset, graciously provided by the authors, by the ideology of the tweet author according to the Southern Poverty Law Center.
    After qualitatively checking sample tweets from each ideology to see how closely they match tenets of white supremacism, we select tweets from the following ideologies: neo-Confederate, neo-Nazi, Ku Klux Klan, racist skinhead, anti-immigration, white nationalist, anti-Semitism, hate music, holocaust identity, Christian Identity.
    44.9\% of tweets were able to be rehydrated from the original set in September 2022.
    Our rehydrated tweets ran from 2009 through 2017.

\begin{figure*}[tb]
    \centering
    \includegraphics[scale=0.4]{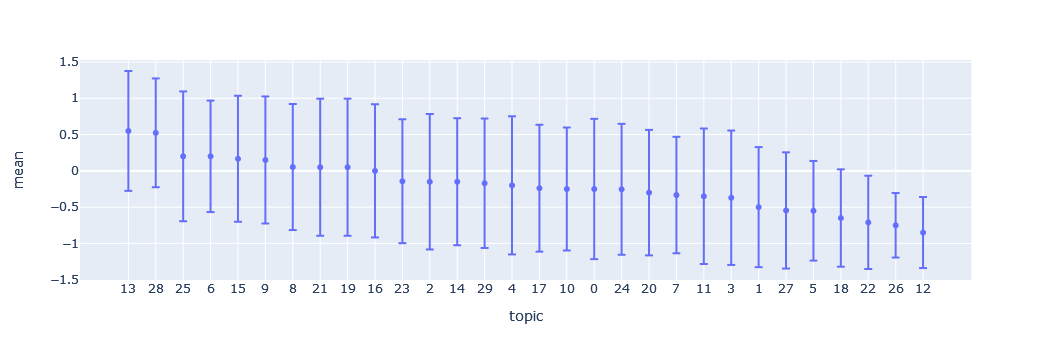}
    \caption{Mean white supremacism annotations by LDA topic in the white supremacist corpus.}
    \label{fig:annotations}
\end{figure*}

\paragraph{Patriot Front data archive:} We select Discord chat posts from servers operated by the white supremacist group, Patriot Front.
These chats were leaked by Unicorn Riot\footnote{\url{https://unicornriot.ninja/2022/patriot-front-fascist-leak-exposes-nationwide-racist-campaigns/}, accessed 11 January 2023}.
After manual inspection for which threads are most ideological, we select the `general' channels from 3 servers: Vanguard America-Patriot Front (2017), Front and Center (2018), MI Goy Scouts Official (2018).

Since chat data may contain names, we remove the top 300 US first names from a 1990 list\footnote{\url{https://namecensus.com/first-names/}, accessed 11 January 2023}.

\paragraph{\citet{calderon_linguistic_2021}:}
We include articles from two white supremacist news websites, the Daily Stormer and American Renaissance, graciously provided by \citet{calderon_linguistic_2021}.
This data contains posts from 2005 through 2017.

\paragraph{\citet{pruden_birds_2022}:}
We include white supremacist books and manifestos collected and provided by \citet{pruden_birds_2022}.
These are: Enoch Powell's ``Rivers of Blood'' speech (1968), Jean Raspail's \textit{Camp of the Saints} (1973, English translation), William Pierce's \textit{The Turner Diaries} (1978), David Lane's ``White Genocide'' manifesto (2012), Anders Breivik manifesto (2011), Renaud Camus' \textit{The Great Replacement} (2012, English translation).
These books and manifestos are split into paragraphs (split at newlines) for experiments.

\begin{figure}[tb]
    \centering
    \resizebox{\columnwidth}{!}{\includegraphics[trim={0 2mm 18mm 14mm},clip]{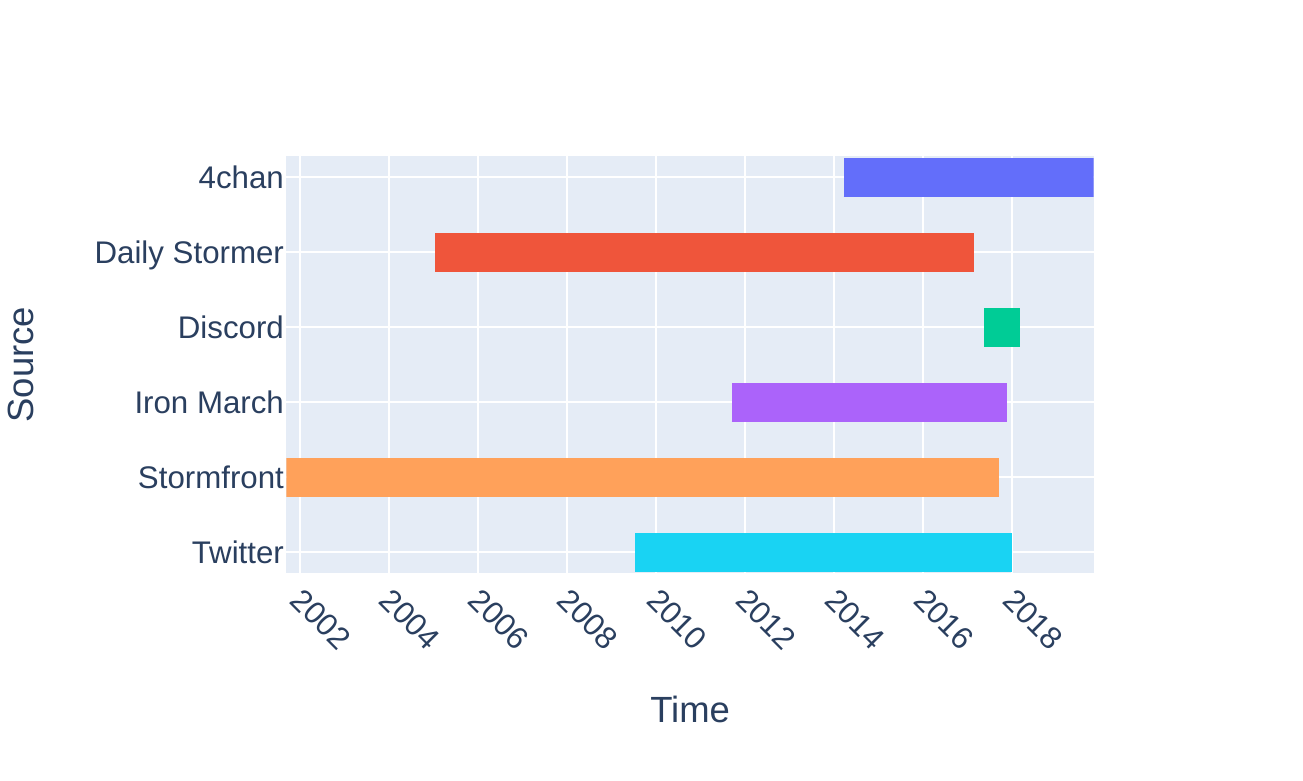}}
    \caption{Time spans of data included in full white supremacist corpus, separated by source. Historical data from \citet{pruden_birds_2022} is excluded.}
    \label{fig:source_spans}
\end{figure}

\begin{figure}[tb]
    \centering
    \resizebox{\columnwidth}{!}{\includegraphics[trim={0 2mm 18mm 14mm},clip]{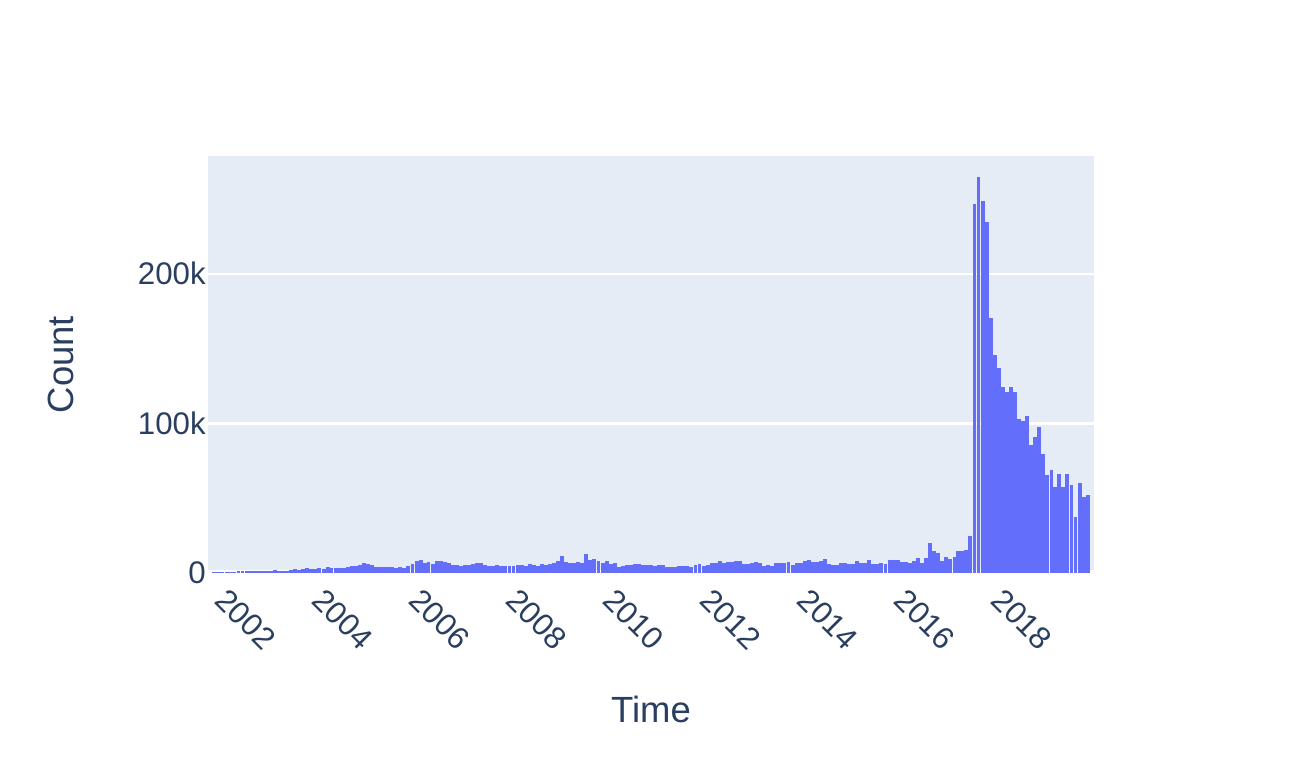}}
    \caption{Histogram of post counts in full white supremacist corpus over time, binned monthly. Historical data from \citet{pruden_birds_2022} is excluded.}
    \label{fig:ws_posts_overtime}
\end{figure}

\paragraph{\citet{elsherief-etal-2021-latent}:}
From this dataset of implicit hate speech tweets\footnote{Available at \url{https://github.com/SALT-NLP/implicit-hate}, accessed 19 January 2023}, we select two portions: 1) tweets labeled for ``white grievance'' by annotators, and 2) when rehydrated, tweets by users identified as holding selected white supremacist ideologies by \citet{qian_hierarchical_2018} (these papers draw on similar datasets).
When we rehydrated these tweets in August 2022, we were only able to access 36.8\%.
Rehydrated tweets spanned from 2009 through 2017.

We lowercase and tokenize all data sources with spaCy 3.1.1 for forum posts and articles, and NLTK's TweetTokenizer~\citep{bird2009natural} for tweets and chat data.

Figure \ref{fig:source_spans} shows the time spans of data from different sources in the full corpus, and Figure \ref{fig:ws_posts_overtime} shows the distribution of posts over time in the dataset.
These figures exclude historical data from \citet{pruden_birds_2022} for readability.

\begin{table*}[tb]
    \centering
    \begin{tabularx}{\textwidth}{|l|X|r|}
\hline
Topic &	Top words & Mean ann. \\
\hline
13 	&	jews jewish jew israel kike anti holocaust kikes zionist goyim & 0.55 \\
28 	& 	white people whites race non black blacks racist hate want 	& 0.52 \\
25 	& eu russia russian europe france french european turks country sweden 	& 0.20 \\
6 	& 	national state people government power nation political socialism society right 	& 	0.20 \\
15 	& 	war hitler germany german did germans nazi world army nazis 	& 0.17 \\
9 	& 	black crime gun kill blacks killed africa rape guns people & 0.15 \\
\hline
    \end{tabularx}
    \caption{LDA topics selected for the white supremacist corpus used in experiments. These are the 6 topics with the highest mean annotation values for white supremacy. \textbf{\textit{Warning: offensive and hateful terms.}}}
    \label{tab:topics}
\end{table*}

\section{Outlier topic removal}
\label{app:lda}
This appendix describes details of removing non-ideological content from our white supremacist corpus.
We run LDA over the full white supremacist corpus and decide on 30 topics after manually inspecting topics for coherence.
We also tried BERTopic~\citep{grootendorst2022bertopic}, but LDA gave a less skewed distribution of documents per topic.

After a brief initial annotation period, one of the authors annotated 20 instances per topic as white supremacist (coded 1), neutral/undecided (0), or not white supremacist (-1).
The criteria was the presence of at least one tenet of white supremacism, described in Section \ref{sec:background}.
Mean distribution of these annotations over topics are presented in \autoref{fig:annotations}.

As can be seen, most topics have mean scores less than 0, i.e., that they contain more posts annotated as neutral or not white supremacist than white supremacist.
This matches results from \citet{rieger_assessing_2021}, who find 24\% of posts in a sample from fringe far-right platforms to be hate speech, high compared to other online spaces but certainly not the majority of posts.
This motivates outlier removal, and we found that removing outlier topics provided an advantage in classification on the evaluation datasets.
Assigning posts to the highest-likelihood topic, we find that filtering to posts within the 6 topics with the highest mean annotations for white supremacy provides the best performance.
As seen in \autoref{fig:annotations}, beyond 6 topics the mean drops to close to a 0 (neutral) rating.
These topics related to antisemitism, anti-Black racism, and discussions of European politics and Nazism.
Top words for these 6 topics are listed in \autoref{tab:topics}.



\section{Evaluation datasets}
\label{app:eval_datasets}
This appendix describes the details of sampling and processing datasets manually annotated for white supremacy used to evaluate classifiers.

\begin{figure}[t]
    \centering
    \resizebox{\columnwidth}{!}{\includegraphics[trim={5mm 10mm 10mm 5mm},clip]{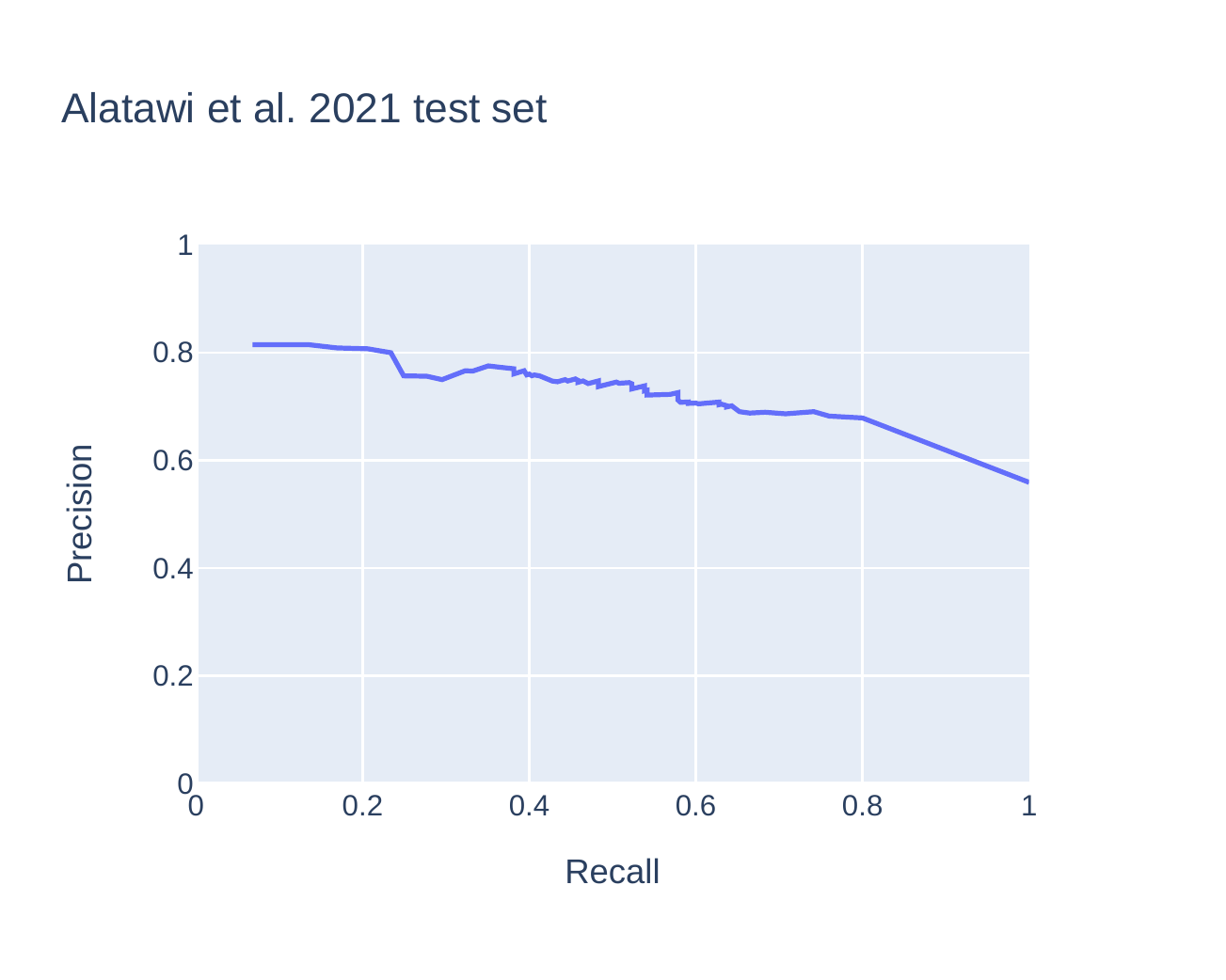}}
    \resizebox{\columnwidth}{!}{\includegraphics[trim={5mm 10mm 10mm 5mm},clip]{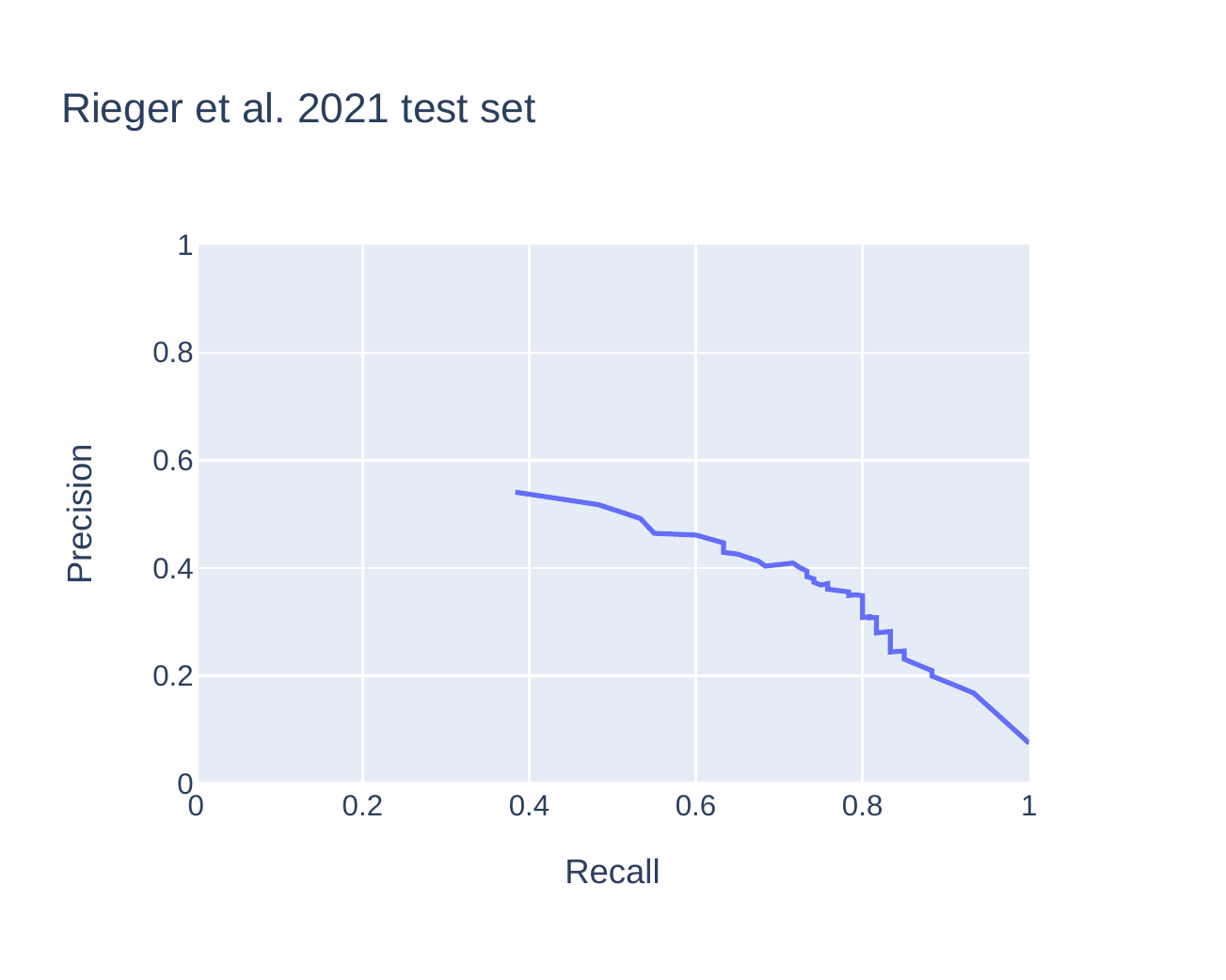}}
    \resizebox{\columnwidth}{!}{\includegraphics[trim={5mm 10mm 10mm 5mm},clip]{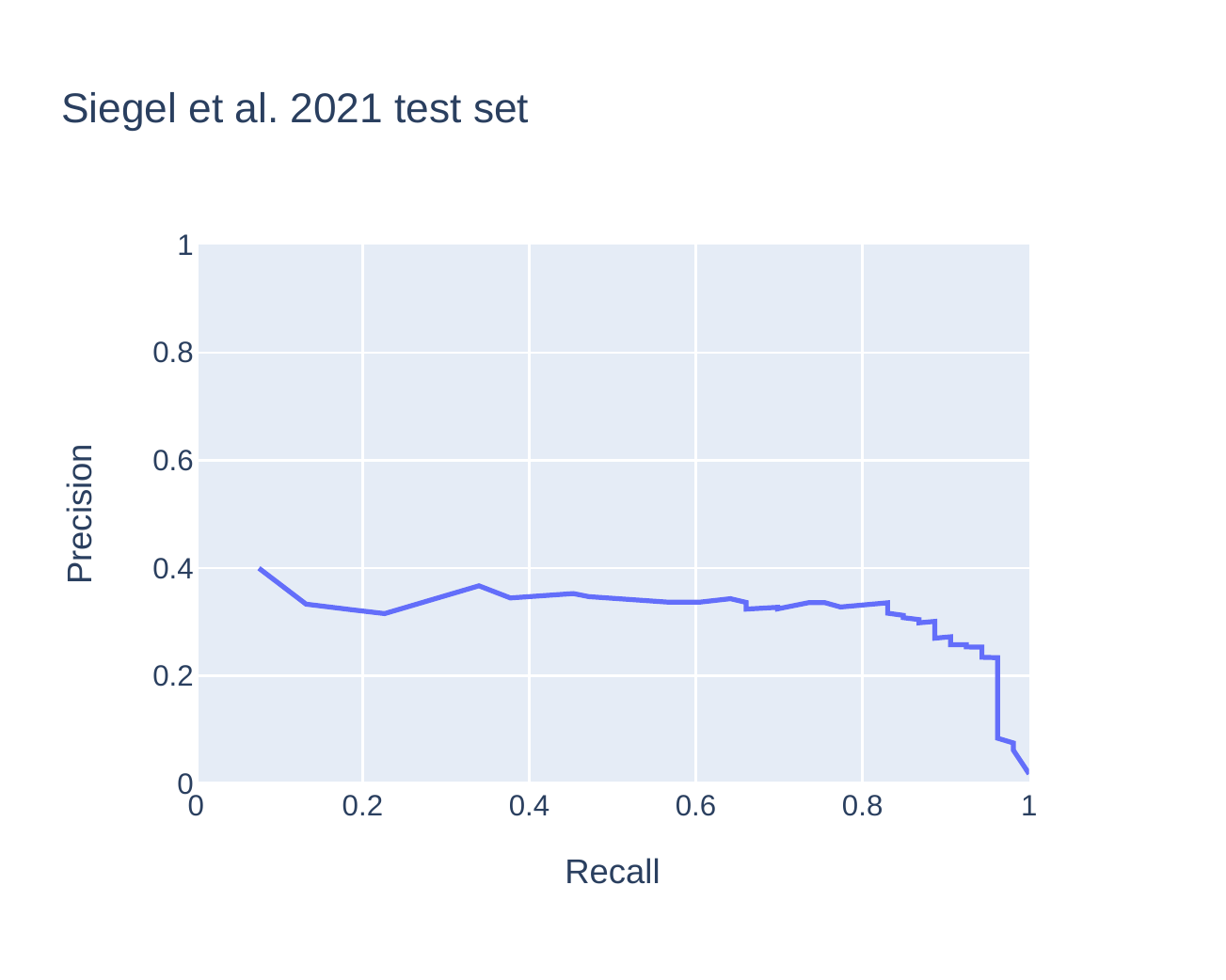}}
    \caption{Precision and recall curves on test splits of evaluation datasets for the best-performing Weak + Annotated model. }
    \label{fig:pr_curves}
\end{figure}

We also present precision and recall curves for our best-performing Weak + Annotated model on evaluation datasets in Figure \ref{fig:pr_curves} for decision thresholds every 0.01 between [0, 1).
Class probabilities were calculated from a softmax over the output class logits.
There is particular room for improvement on precision for \citet{rieger_assessing_2021} and \citet{siegel_trumping_2021} datasets.

\paragraph{\citet{alatawi_detecting_2021}:}
From the full annotated dataset of tweets from \citet{alatawi_detecting_2021}\footnote{Accessed from \url{https://github.com/Hind-Saleh-Alatawi/WhiteSupremacistDataset} on 11 January 2023.}, we choose the combined annotator labels for white supremacy as the label of white supremacy or not.

\paragraph{\citet{rieger_assessing_2021}:}
This dataset, provided by the authors, contains posts on fringe platforms (4chan /pol/, 8chan /pol/, and r/the\_Donald) annotated for many aspects of hate speech, including white supremacist ideology.
We sample examples labeled for `white supremacy/white ethnostate' or `National Socialist' ideology as examples of white supremacy.
For negative examples, we sample posts that are not labeled as white supremacist or as hate speech for negative examples, since their definition of white supremacy is more restrictive
Specifically, we sample posts not labeled for `white supremacy/white ethnostate', `National Socialist', `general insult', `personal insult' or `violence'.
Direct requests for this dataset to the authors.

\paragraph{\citet{siegel_trumping_2021}:}
We use training data from \citet{siegel_trumping_2021}, provided by the authors.
From lists of tweets annotated for white nationalism and hate speech, we select those marked as positive for white nationalism and as negative examples, those annotated as neither white nationalism nor hate speech.
Requests for this dataset should be directed to the authors.

\end{document}

%% file: tikz/venn_diagram.tex
\def\firstcircle{(0,0) circle (1.9cm)}
\def\secondcircle{(0:1.9cm) circle (1.9cm)}

\colorlet{circle edge}{black!50}
\colorlet{circle area}{black!20}

\tikzset{filled/.style={fill=circle area, draw=circle edge, thick},
	outline/.style={draw=circle edge, thick}}
\tikzset{fillred/.style={fill=black!5, draw=circle edge, thick},
	outline/.style={draw=circle edge, thick}}

\begin{figure}[tb]
    \centering
    \begin{tikzpicture}
    	\begin{scope}
        	\clip \firstcircle;
        	\fill[filled] \secondcircle;
    	\end{scope}
    	\draw[fillred, even odd rule] \firstcircle node {}
                                 	\secondcircle node {};
    	\draw[outline] \firstcircle node[text=black,left=-0.5mm,align=center] {white\\supremacy};
    	\draw[outline] \secondcircle node[text=black,right=2mm,align=center] {hate\\speech};
    \end{tikzpicture}
    \caption{Conceptualization of the relationship between hate speech and white supremacism used in this paper. Much of white supremacist language includes text that would be considered hate speech, i.e. attacks against those with marginalized identities. However, we also aim to capture text that expresses white supremacist ideology without direct hate speech, such as the glorification of Nazism. Finally, some hate speech would not fit as expressing a white supremacist ideology, such as antisemitism within a Black Nationalist context.}
    \label{fig:venn_diagram}
\end{figure}